\documentclass[conference]{IEEEtran}
\usepackage[utf8]{inputenc}
\usepackage{enumerate}
\usepackage{multirow}
\usepackage{array}
\usepackage{adjustbox}
\usepackage{makecell}
\usepackage{tabularx}
\usepackage{float}
\usepackage{url}
\usepackage{amsfonts, amsmath, amsthm, amssymb}
\usepackage{xcolor}

\usepackage{cite}
\def\BibTeX{{\rm B\kern-.05em{\sc i\kern-.025em b}\kern-.08em
    T\kern-.1667em\lower.7ex\hbox{E}\kern-.125emX}}

\ifCLASSINFOpdf
 
\else
 
\fi

\begin{document}

\title{A Electric Network Reconfiguration Strategy with Case-Based Reasoning for the Smart Grid}

\author{\IEEEauthorblockN{Flávio G. Calhau}
\IEEEauthorblockA{Universidade Federal da Bahia - UFBA\\
Email: fgcalhau@gmail.com}
\and
\IEEEauthorblockN{Joberto S. B. Martins}
\IEEEauthorblockA{Universidade Salvador - UNIFACS\\
Email: joberto.martins@gmail.com}
}

\maketitle

\begin{abstract}
The complexity, heterogeneity and scale of electrical networks have grown far beyond the limits of exclusively human-based management at the Smart Grid (SG). Likewise, researchers cogitate the use of artificial intelligence and heuristics techniques to create cognitive and autonomic management tools that aim better assist and enhance SG management processes like in the grid reconfiguration. The development of self-healing management approaches towards a cognitive and autonomic distribution power network reconfiguration is a scenario in which the scalability and on-the-fly computation are issues. This paper proposes the use of Case-Based Reasoning (CBR) coupled with the HATSGA algorithm for the fast reconfiguration of large distribution power networks. The suitability and the scalability of the CBR-based reconfiguration strategy using HATSGA algorithm are evaluated. The evaluation indicates that the adopted HATSGA algorithm computes new reconfiguration topologies with a feasible computational time for large networks. The CBR strategy looks for managerial acceptable reconfiguration solutions at the CBR database and, as such, contributes to reduce the required number of reconfiguration computation using HATSGA. This suggests CBR can be applied with a fast reconfiguration algorithm resulting in more efficient, dynamic and cognitive grid recovery strategy.

%EVENTUALLY TO BE USED: SGs do require in some scenarios scalable on-the-fly computation for the reconfiguration of large distribution power networks.
%Smart City is one scenario in which scalable on-the-fly computation is required for fast network reconfiguration.

\end{abstract}


\IEEEpeerreviewmaketitle

\section{Introduction}

Smart Grid represents a modern vision of a dynamic electricity grid, in which electricity and related information flow together in real time, allowing near-zero economic losses in the event of outages and power quality disturbances. All of this being supported by a new energy infrastructure built on top of communication channels, distributed intelligence and possibly clean power \cite{Sarvapali_2012}.

Current electricity grids are highly heterogeneous, have to deal with an exponential growth in the number of users, are highly dynamic in terms of user's demands and are subject to failure \cite{mahmoud_cognitive_2007}. Either in case of failure or to allow maintenance maneuver and optimization, the network must be reconfigured as rapidly as possible.

%Outages on power supply are inevitable, whether for system components preventive maintenance interventions or operation of a protective device in the presence of a failure, among other possibilities. A maneuver plan must exist to deal with the effects resulting from interruptions, to minimize the affected area and restore power supply of such areas as soon as possible.

%It is also a fact that the research and developing communities have struggled in the last years to provide adaptable management and recovery solutions towards autonomic networks. Autonomic solutions aim to reduce human intervention in complex management tasks and engineer accurate knowledge, possibly on-the-fly, to better support the management process  \cite{bezerra_network_2014}.

%The computation of distribution network reconfiguration consists basically in finding, among all possible combinations, a set of configurations that (i) minimize the number of switching operations while keeping the radial structure of the network, (ii) reduce the energy loss and (iii) reduce the number of consumers without electricity supply. In addition, it is necessary to safeguard voltage levels restrictions at the point of load, maximize current flow in the branches and safeguard lines flow capacity and nominal power of the transformers \cite{KIRAN_2013}.

The use of artificial intelligence to create cognitive distribution power network reconfiguration management tools that aim better assist and enhance the SG distribution network reconfiguration processes is an important research issue \cite{colak_survey_2016}.

In addition to the artificial intelligence component, fast algorithms are also necessary to compute new distribution network reconfiguration. In effect, the overall distribution power network reconfiguration solution must scale to be adequate for on-the-fly utilization in large grid deployments, like the ones existing in Smart Cities.

%he Smart City is one scenario in which scalable on-the-fly computation is required for fast network reconfiguration. The adoption of self-healing management approaches towards a cognitive and autonomic distribution power network reconfiguration is another scenario in which the scalability and on-the-fly computation are issues. 

This paper proposes the use of Case-Based Reasoning (CBR) coupled with the HATSGA algorithm for achieving fast reconfiguration of large distribution power networks. The motivation is to develop a CBR-based framework with cognitive self-healing characteristics for the distribution network aiming the reduction of human intervention in the recovery process.

%Scalability is an essential factor for performance evaluation and optimization of parallel and distributed systems \cite{M.Wu:2005}. It has been widely used for describing system's capability to handle a large amount of work and how the system size and problem size will influence the performance of algorithms.

%We propose in this paper a network recovery strategy for Smart Grid based on Case-Based Reasoning for distribution network reconfiguration problem that ...for energy loss minimization.

The paper is structured with section 2 initially presenting the related research. Section  III describes the conceptual frameworks adopted (CBR-SGRec). The HATSGA  algorithm \cite{FGCADVANCE:2015}, used for the reconfiguration computation, is presented and evaluated in sections III and IV. The cognitive CBR-based approach is presented and evaluated in sections VI and VII. Final considerations are presented in section VIII.

\section{Related Work}

In recent years, significant research has been done to minimize power loss in the process of reconfiguring distribution power systems. The reconfiguring of distribution power system has a combinatorial nature  and the search of the best computational time for supporting the decision-making process in real time has been focused on \cite{Thomas}, \cite{Haughton}, \cite{Merdan}.

Tabu search algorithm has been used for several combinatorial optimization solutions \cite{R.Cherkaoui} \cite{Thakur} and \cite{abdelaziz2010distribution}. In \cite{Thakur} Tabu search is used as a meta-heuristic method for network reconfiguration problem in radial distribution systems. In \cite{abdelaziz2010distribution} is proposed a method of network reconfiguration using a modified Tabu search algorithm focusing on reducing keys opening and closing and minimizing the loss. The HATSGA algorithm proposes an enhanced Tabu list to compute only more relevant data and reduce the computation time \cite{FGCADVANCE:2015}.

In \cite{Jakus__Sarajcev_Vasilj_2017} the authors implement an algorithm for network reconfiguration for a realistic distribution network based on a genetic algorithm (GA), taking as objective power loss minimization and load balancing index. HATSGA uses a distinct genetic algorithmic approach by using elitism to choose among potential solutions. 

%In terms of having a macro view of available techniques for network reconfiguration, the Figure \ref{fig:Mindmap} shows the main approaches used. Methods based on hybrid algorithms are indicated as having the potential to significantly reduce the time to compute solutions \cite{KIRAN_2013}.

%%=== Figura 01 ====
%\begin{figure}[!h]
%  \begin{centering}
%  \centering
%    \includegraphics[width=0.5\textwidth]{_Thesis_paper/imagens/MindMaps-eps-converted-to.pdf}
%  \end{centering}
%  \caption{Reconfiguration Problem Approaches}
%  \label{fig:Mindmap}
%\end{figure}

%To the best of our knowledge, none of these network reconfiguration approaches has talked yet the using Machine Learning using CBR applied for the network reconfiguration problem. \textcolor{red}{(NOT VALID ANYMORE; THER ARE A NUMBER OF APPROACHES FOR THAT FLAVIO)}

In general, these proposals typically address a specific existing heuristic or specific algorithms to implement reconfiguration process of electrical distribution network. They do not take into account a scalability analysis and the computational time to find solutions.

%The scalability issue must be considered by current approaches solutions under risk of such solutions fail to fit into different reconfiguration scenarios. However, it does not address issues related to the run time in solutions search and do not address flexibility and operational performance using machine learning. 

%\section{Network Reconfiguration with Case-Based Reasoning - Basic Approach and Issues}
\section{THE CONCEPTUAL CBR-SGRec FRAMEWORK}

\begin{figure}[ht]
%  \begin{centering}
  \centering
    \includegraphics[width=0.53\textwidth]{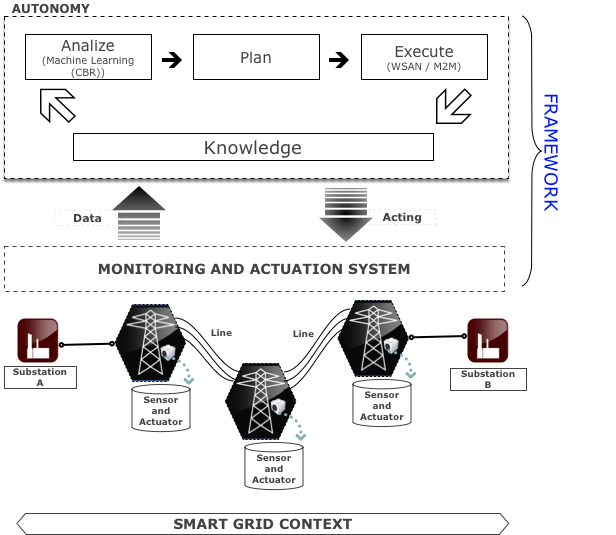}
%  \end{centering}
  \caption{The conceptual CBR-SGRec framework}
  \label{fig:Arcabouco_EN}
\end{figure}

The conceptual CBR-based Smart Grid network recovery framework (CBR-SGRec) is illustrated in Figure \ref{fig:Arcabouco_EN}.

%This section defines the framework with autonomic characteristics presenting its structure for monitoring and solving the network reconfiguration problem.

%The complexity, heterogeneity, and scale of networks have grown far beyond the limits of manual administration. Furthermore, the main cause of outages in many network environments is human error \cite {Ayoubi_Limam__2018}. 

The basic CBR-SGRec framework componets are:
\begin{itemize}
    \item The smart grid distribution network;
    \item A monitoring system collecting grid operation parameters;
    \item A knowledge plan; and
    \item An actuation system capable to deploy new reconfiguration topologies on the the network.
\end{itemize}

%Figure \ref{fig:Arcabouco_EN}, with monitoring and actuation capabilities with a knowledge plan that, in short, addresses the problem of reconfiguration of electrical networks using autonomic characteristics. 

%The framework, Figure \ref{fig:Arcabouco_EN}, with monitoring and actuation capabilities with a knowledge plan that, in short, addresses the problem of reconfiguration of electrical networks using autonomic characteristics.

%%=== Figura 02 ====

The CBR-SGRec framework knowledge plane includes the basic elements of the CBR strategy:
\begin{itemize}
    \item A knowledge database containing possible network reconfiguration solutions;
    \item The CBR engine analyzing, planning and acting on behalf of the network reconfiguration process; and
    \item The network reconfiguration algorithm to be called whenever required. 
\end{itemize}

%The basic research scenario proposed is the electrical systems aiming at the feasibility of solutions optimized for the smart cities and for the systems of management of distribution of energy in the SG context.

%The communication system in the smart grid solution is, as a rule, a comprehensive solution from the technological point of view. As an example, the communication system for the smart grid has distinct elements that include specific LAN solutions for the internal communication between the control elements in the substations (IEDs, actuators, others), long distance communication systems supporting efficient and reliable power transmission, support for monitoring and collecting data from different types of systems sensors.

In relation to the classical smart grid reconfiguration management strategy,  the CBR-SGRec approach improves current solutions by using machine-learning techniques coupled with an efficient reconfiguration algorithm to find qualified network reconfiguration solutions with reduced computational time.

The analysis and plan function lists symptoms in order to diagnose the problem. Once the problem has been determined, policies are accessed to direct the actions that will be taken (HATSGA and CBR), indicating an appropriate solution to the problem to the manager so that it can aid in the decision making in a faster and efficient way. Thus generating an execution plan. The execution plan receives an indication of action and applies it.

.

\section{HATSGA Algorithm}

HATSGA is an algorithm aimed to compute power distribution reconfiguration solutions in the Smart Grid context \cite{FGCADVANCE:2015}. HATSGA uses graph theory with language “R” to model the distribution network. HATSGA uses only radial topologies and reduces the search space to minimize power flow evaluation and to reduce the computational algorithm effort required.

HATSGA's strategy minimizes the search space solution by eliminating topology configurations that do not comply with constraints criteria like radial configuration, voltage profile and system loss.

HATSGA uses elitism, a technique inspired by evolutionary biology and natural selection \cite{Zuben2000}. Elitism is used by HATSGA to select potential topologies that can be used to compute new reconfiguration solutions by selecting electrical parameters that may result in the computation of new minimal solutions for power loss.

The Tabu Search is a technique that uses a list to store found solutions that should not be considered in the computation (forbidden). HATSGA uses a modified Tabu Search algorithm. This is achieved by introducing to the conventional “tabu list” a set of associated parameters. In effect, the HATSGA tabu list is a bi-dimensional matrix composed by a list of open switches and the power loss computed for all the resulting topologies. Another aspect differentiating the conventional tabu list from the HATSGA`s one is that the list is always kept during computation to allow optimization in terms of the computational time.

The summary of HATSGA algorithm phases are as follows (Figure \ref{fig:flowchartmain}):

	%=== Figura 5 ====
\begin{figure}[ht]
\centering
\includegraphics[width=0.47\textwidth]{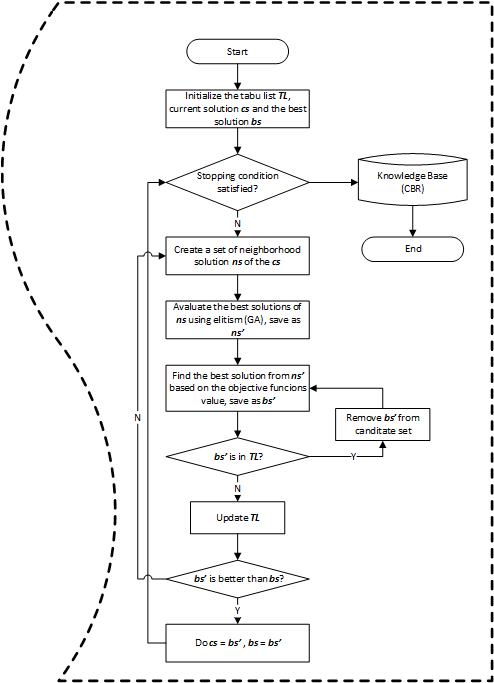}
\caption{HATSGA's execution flow}
\label{fig:flowchartmain}
\end{figure}

\begin{enumerate}
	\item  HATSGA generates a radial topology network as an initial topology using the minimum spanning tree  (\textit{cs}), calculate the power loss (\textit{bs}) for this topology through the power flow calculation based on Newton-Raphson and build a tabu list  (\textit{TL}) with the initial configuration (status of open edges).
	\item From (\textit{cs}), all the open switches (edges) are stored in a vector (\textit{ns}). For each switch in ns, changed the status ``closed'', creating a loop in the current topology (\textit{cs}).
	\item Use elitism (where n of the best candidates in each generation are taken to the next generation) to select the topologies that have a greater probability of success to be part of the solution and these are in store vector \textit{ns'}. 
	\item for each switch in ns', the switch is open undoing the loop. The loss power of the new topology is calculated. If the new topology is not yet stored in the tabu list (\textit{TL}), it is added.
	\item If the power loss \textit{bs'} of the new computed topology is lower than the previously stored (\textit{bs}), the best solution is updated.
	\item After checking all existing sectorial loops, power loss  \textit{bs} of the solution is updated with the best topology.
\end{enumerate}

\section{HATSGA Evaluation}

 The objective of the HATSGA evaluation is to verify its capability to scale for large networks computing the network reconfiguration within an acceptable time. The aspects evaluated are:

\begin{enumerate}
	\item Verification of HATSGA capability to commute with a large amount of switches and how the system size will influence its performance.
	\item The computational time required to compute solutions.
\end{enumerate}

These algorithm characteristics are fundamental requirements to compute an intelligent and on-the-fly network reconfiguration.

HATSGA capability to scale will be evaluated by using the IEEE N-Bus test scenarios with an increasing number of buses and switches. The test scenarios used were the IEEE 14-Bus, IEEE 30-Bus, IEEE 57-Bus, IEEE 118-Bus and IEEE 300-Bus tests system \cite{Testsystem:2017}. 

Table \ref{tab:scalabilityscenarios} presents the number of buses, switches and topologies that are manipulated by the algorithm for the distribution network reconfiguration computation. The search space increase nearly exponentially from the 14-Bus to the 300-Bus and this requires an algorithm strategy to maintain an acceptable computational time.

\begin{table}[H]
\caption{HATSGA Scalability IEEE Test System Scenarios}
\label{tab:scalabilityscenarios}
\begin{center}
%\begin{adjustbox}{width=0.49\textwidth} 
\begin{tabular}{|c|c|c|c|} \hline
\thead{Topology \\ IEEE}     & \thead{Buses} & \thead{Switches} & \thead{Numbers  \\ Spanning \\ Tree} \\ \hline
14-Bus  & 14    & 20   & 3909                 \\ \hline
30-Bus  & 30    & 41   & 7824000              \\ \hline
57-Bus  & 57    & 80   & $2.193e+20$          \\ \hline
118-Bus & 118   & 186  & $2.159e+41$          \\ \hline
300-Bus & 300   & 411  & $2.366e+64$          \\ \hline       
\end{tabular}
%\end{adjustbox}
\end{center}

\end{table}

The quality of the solution is another aspect of the computed reconfiguration. In our case, it is determined by defining limits for the voltage profile and power system loss parameters.

\subsection{HATSGA Scalability Test Results}

The simulation run used a Macbook with an Intel core i7 (dual core) 2.6 Ghz CPU and 8 GB RAM using MacOS Sierra (version 10.12.5). The algorithm execution time was computed by the function proc.time available with the “R” programming environment. This function determines how much computational time the HATSGA code consumes.

Table \ref{tab:scalabilityresults} presents HATSGA execution time for IEEE test bus systems. As far as our knowledge is concerned, the literature only presents the computational time requires for IEEE 14-Bus test minimum power loss. HATSGA algorithm, as a figure of merit, gets minimum power loss results that is equivalent to the best result obtained by this algorithm described in \cite{FGCADVANCE:2017} and \cite{NACHIMUTHU}.

\begin{table}[H]
\caption{HATSGA Scalability Results on IEEE Test System}
\label{tab:scalabilityresults}
\begin{adjustbox}{width= 0.49\textwidth} 
\begin{tabular}{|c|c|c|c|c|c|} \hline
\thead{Topology \\ IEEE}     & \thead{Mean Time\\ (in sec)} & \thead{Standard \\ deviation} & CI (95\%) \\ \hline
14-Bus  & 1,82    & 0,043   & [1,7963 : 1,8367]          \\ \hline
30-Bus  & 12,57   & 1,336   & [11,714 : 13,423]          \\ \hline
57-Bus  & 16,94   & 0,946   & [16,347 : 17,533]          \\ \hline
118-Bus & 96,04   & 14,170  & [87,202 : 104,887]         \\ \hline
300-Bus & 1380,68 & 58,409  & [1344,37 : 1416,98]        \\ \hline       
\end{tabular}
\end{adjustbox}
\end{table}

\begin{figure}[ht]
\centering
\includegraphics[width=0.5\textwidth]{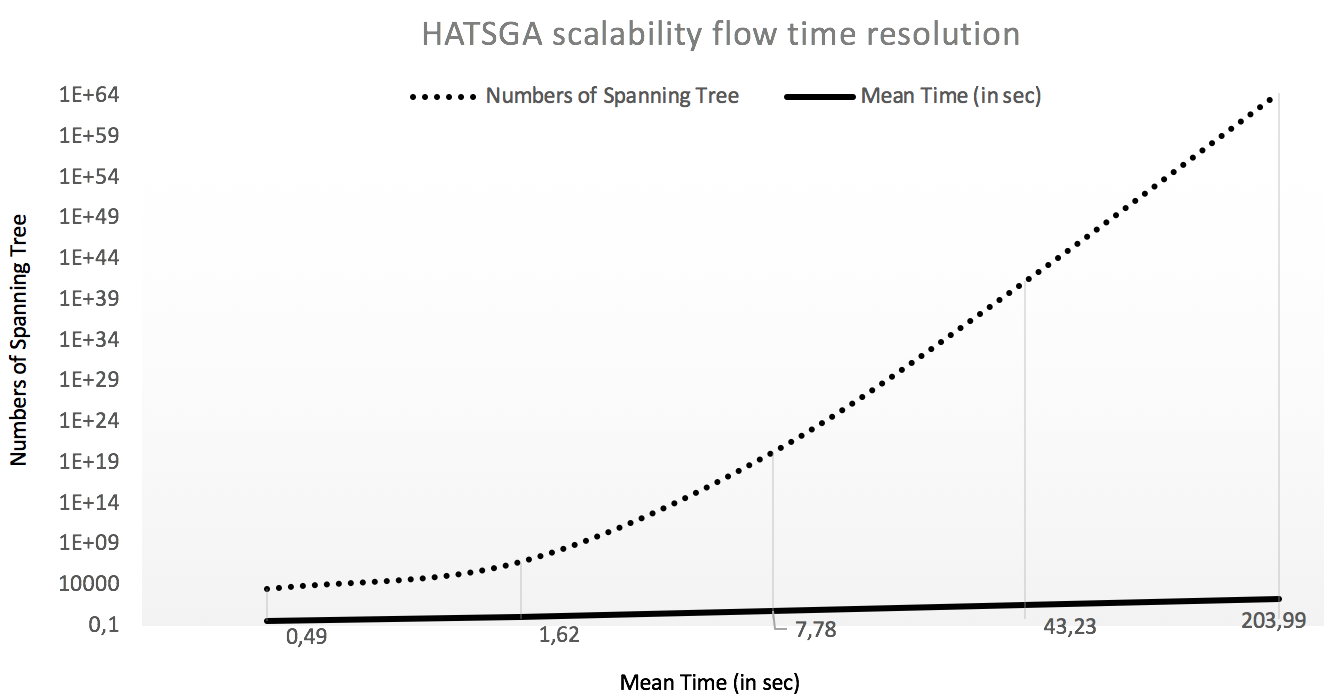}
\caption{HATSGA scalable behavior with network complexity}
\label{fig:flowtimesolution}
\end{figure}

The Figure \ref{fig:flowtimesolution} illustrates the scalability of HATSGA algorithm, by indicating the required reconfiguration computational time for large power networks. It shows a linear increase for the solution search time in contrast to the exponential growth of the topology complexity. 

	%=== Figura 7 ====

This result suggests the viability of using the HATSGA algorithm for on-the-fly network reconfiguration and to support the cognitive CBR-SGRec framework approach for large distribution power networks.

\section{THE CBR-SGREC KNOWLEDGE PLAN WITH CBR}

Case-Based Reasoning (CBR) is a machine learning technique for problem solving that solves new problems using the experience acquired with previous cases \cite{aamodt_case-based_1994}. CBR functions  as  a cognitive  model  that  allows to imitate  humans  to solve real problems by remembering previously solved cases (problems) that are used to suggest a solution for novel but similar situation.

CBR module allows the CBR-SGRec framework to accelerate the proposition of solutions for the network reconfiguration problem based on the stored cases.

\subsection{CBR Modelling for the Network Reconfiguration Problem}

In CBR one case is a stored pair with a problem and a solution used to solve this problem in the past. Thus, the cases (\textit{c\textsubscript{i}}) are composed of problems encountered in the network  (\textit{p\textsubscript{i}}) and the solution to such problems (\textit{s\textsubscript{i}}):

\begin{equation} 
\label{eq:case_model}
\textit{c\textsubscript{i}} = \{\textit{p\textsubscript{i}}, \textit{s\textsubscript{i}}\}
\end{equation}

The generic CBR case definition (\ref{eq:case_model}) has been extended for CBR-SGRec to ensure more efficiency and flexibility in the CBR database search (\ref{eq:case_model_extended}). Flexibility concerns the modelling of problems that can be caused in different contexts. As an example, a bus may fail and cause different reactions depending on the current state of the network (interrupt different areas depending on priority). The efficiency is achieved by having and strategy to reduce the number of cases in the database.

CBR case (\textit{c\textsubscript{i}}) definition for CBR-SGRec  is as follows:

\begin{equation} 
\label{eq:case_model_extended}
\textit{c\textsubscript{i}} = \{\textit{sn\textsubscript{i}}, \textit{p\textsubscript{i}}, \textit{s\textsubscript{i}}, \textit{ls\textsubscript{i}}, \textit{qs\textsubscript{i}}, \textit{ns\textsubscript{i}}\}
\end{equation}

Where,

\begin{itemize}
		\item \textit{sn\textsubscript{i}} – represents the current state of network;
		\item \textit{p\textsubscript{i}} – corresponds to problems encountered in the network that cause instability in the operation. In network reconfiguration, this can be: power failure, imbalance of electric power and maintenance of components, among others;
		\item \textit{s\textsubscript{i}} – represents the solution to the problem;
		\item \textit{ls\textsubscript{i}} – represents the power loss of the topology;
		\item \textit{qs\textsubscript{i}} – represents the proposed solution quality. As an example, voltage at any given bus must not exceed 5\% of the nominal value;
		\item \textit{ns\textsubscript{i}} – indicates the number of occurrences of a case \textit{c\textsubscript{i}}.
\end{itemize}

\begin{figure}[ht]
\centering
\includegraphics[width=0.47\textwidth]{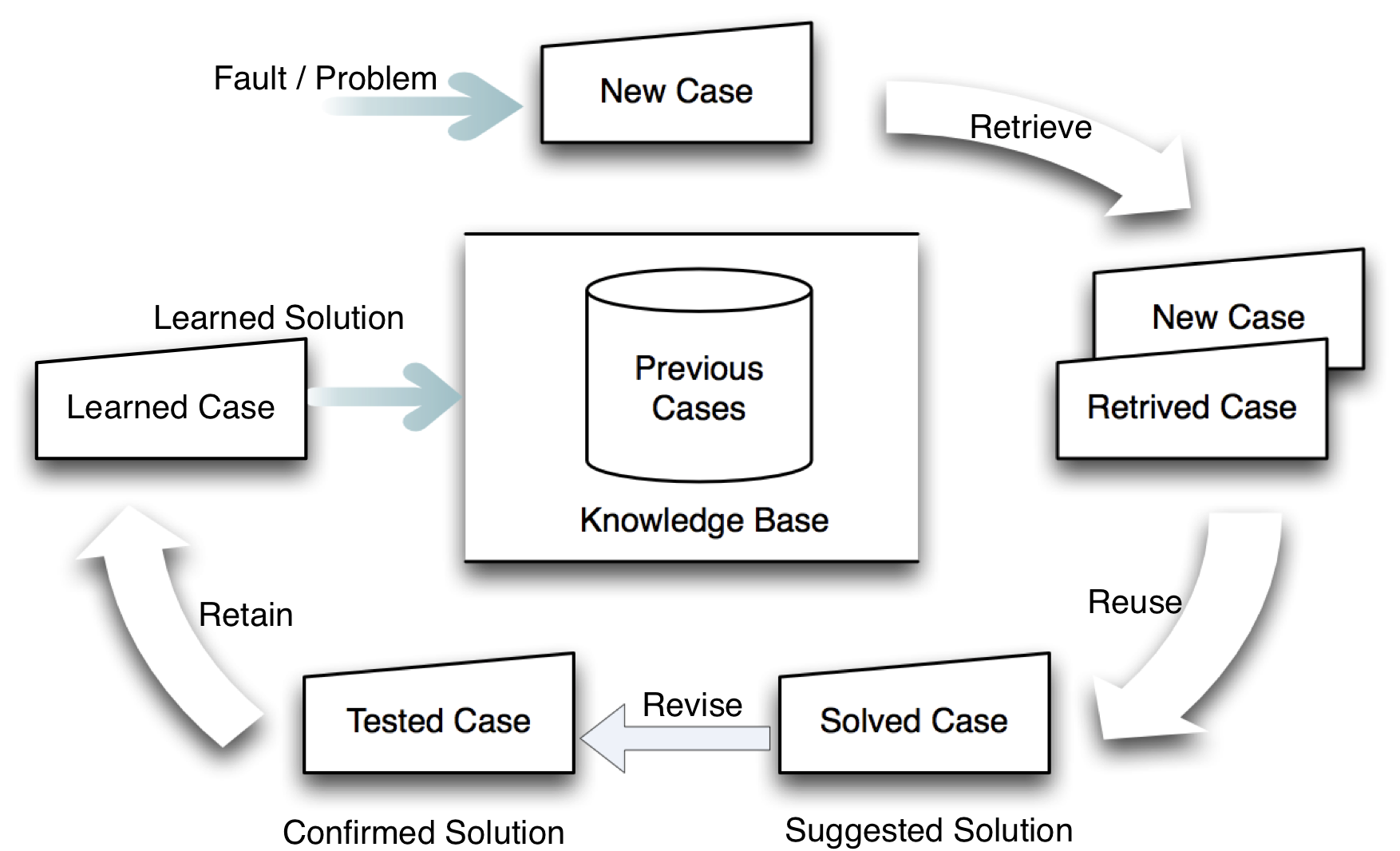}
\caption{Case-Based Reasoning Cycle – Adapted from  \cite{aamodt_case-based_1994}}
\label{fig:CBR_Cycle}
\end{figure}

CBR workflow in the CBR-SGRec framework follows the classical 4R sequence (Figure \ref{fig:CBR_Cycle}) \cite{aamodt_case-based_1994}. 

The process is triggered by an existing problem (new case). A similar case is searched in the CBR database using an appropriate similarity function and an adaptation is done if necessary. The solution is applied and, later validated for retention in the solutions CBR database.   

	%=== Figura 6 ====

\subsection{CBR Similarity}

One of the reasons for the use of similarity follows the hypothesis that similar problems must have similar solutions \cite{Wangenheim_2003}. There are many ways of evaluating similarity with different approaches for different case representations \cite{FGCADVANCE:2019}.

Sequentially processing all cases in memory for evaluating similarity has complexity \textit{O(n)}, where \textit{n} is the number of cases.  If \textit{n} is very large,  can affect the computational response time in the search for solution. Aiming to minimize processing overhead and reduce case retrieval time to achieve efficiency, the framework performs maintenance on the database removing cases not recently used. This is done using the number of occurrences of the extended case definition (\ref{eq:case_model_extended}).

 In this work, a local similarity is defined for each attribute and a global similarity is computes as a weighted average of the local similarities.

CBR-SGRec framework calculates the local similarity between numerical attributes  using  the  distance  between  them,  calculated  by  the module  of  the  difference  between  the attributes  as  indicated  in equation \ref{eq:local_similarity}\cite{Shiu_Pal_2004}.

\begin{equation} 
\label{eq:local_similarity}
\textit{f}(\textit{T\textsubscript{i}}, \textit{C\textsubscript{i}}) = \frac{1 - \left | Ci - Ti \right |}{\left ( max - min \right )}
\end{equation}

The global similarity is defined as a computed  mean of the  distinct local similarities for the attributes. It is based on weighted Euclidean distance (\ref{eq:global_similarity}).

\begin{equation} 
\label{eq:global_similarity}
Sim\left ( Ti,Ci \right ) =
\sqrt{\sum_{i = 1}^{n} \textit{w\textsubscript{i}}\left ( \textit{t\textsubscript{i}} - \textit{c\textsubscript{i}} \right )^2}
\end{equation}

Where,

\begin{itemize}
		\item \textit{T\textsubscript{i}} is the input case.
		\item \textit{S\textsubscript{i}} is the case stored in CBR database.
		\item \textit{n} is the attribute number of each case.
		\item \textit{t\textsubscript{i}} is attribute-value pairs representing the indices \textit{T} = (\textit{t\textsubscript{1}, \textit{t\textsubscript{2},...,\textit{t\textsubscript{n})}}}.
		\item \textit{c\textsubscript{i}} is attribute-value pairs representing the indices \textit{C} = (\textit{c\textsubscript{1}, \textit{c\textsubscript{2},...,\textit{c\textsubscript{n})}}}.
		\item \textit{w} is the weight given to the attribute \textit{i}
\end{itemize}

\section{CBR SUPPORTING THE NETWORK RECONFIGURATION STRATEGY}

The CBR-SGRec framework process two typical and distinct situations: (a) a fault triggers the reconfiguration or subsidize the manager in the decision making concerning the network topology and (b) a reconfiguration is triggered to improve the balance or the quality of the network.

The  CBR-SGRec knowledge plan starts  with  the  CBR module receiving an “alert” indicating a fault or that the limits previously set by the network manager are not being obeyed. The CBR module gets from monitoring module all values (attributes) of the network that describe its current state. This set of information is named "Current Case" or “case” for short. Having the "case" the CBR module triggers the first step of the CBR 4R cycle (Figure \ref{fig:CBR_Cycle}). 

At  this  stage, the CBR module checks the existence or not of a stored case with similar characteristics in relation to the actual case that generated the alert. This phase returns the most significant cases (similar ones) towards a possible solution for the current case. If a similar case is found, the next phase (reuse) is called.  

If there is no similar case, the decision plan triggering the reconfiguration algorithm HATSGA to compute a solution to the  current  problem.  The  current  case,  along  with  HATSGA's solution are compiled and become a "New Case" that is sent for execution (using or not manager's approval) and following that is checked for retention. 

As indicated, the decision plan can execute a new case without the intervention  of the manager. With this option, an arbitrary solution is attributed to the current problem and checked for suitability.

\subsection{CBR Module Proof of Concept}

The CBR module was implemented using the Protégé \cite{protege} and MyCBR \cite{MyCBR} tools. Protégé is a free, open-source ontology editor and framework for building CBR systems. MyCBR is an open-source similarity-based retrieval tool and software development kit (SDK). The network topology used in the proof-of-concept was the IEEE-14Bus.

\begin{figure}[ht]
\centering
\includegraphics[width=0.47\textwidth]{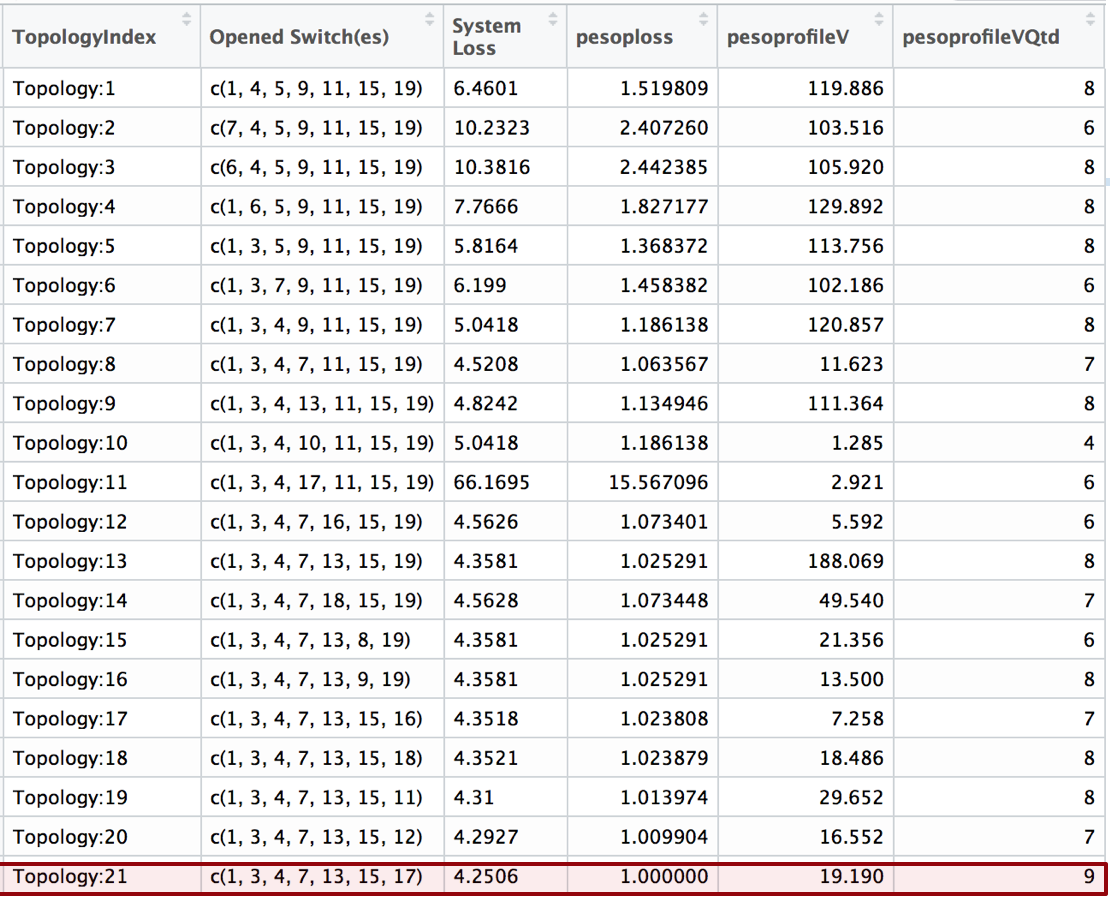}
\caption{Example of CBR database for IEEE 14 Bus System}
\label{fig:CBR_Base_14Bus}
\end{figure}

As part of the CBR module proof-of-concept, the case database was populated with solutions obtained using HATSGA, as show in Figure \ref{fig:CBR_Base_14Bus}.

	%=== Figura 8 ====

The Protégé configured parameters to evaluate similarity were "pesoploss" used to show the system loss ratio topology similarity, the "pesoprofileV" that represents the accumulate load-flow in the system voltage profile and "pesoprofileVQtd" that represents the number of buses that exceeded the limit of the nominal value. A similarity threshold and similarity parameter priority may also be defined for the CBR module.

Figure \ref{fig:CBR_Base_14Bus} illustrate the current topology (Topology21) and other topologies with less accurate operational parameters from the network management perspective.

In terms of the proof of concept, a fault is simulated eliminating buses 9 and 11 at the current topology. A global similarity of 92\% is defined, no priority is assigned to the similarity parameters and an alert is sent to the CBR-SGRec framework.

\begin{figure}[ht]
\centering
\includegraphics[width=0.47\textwidth]{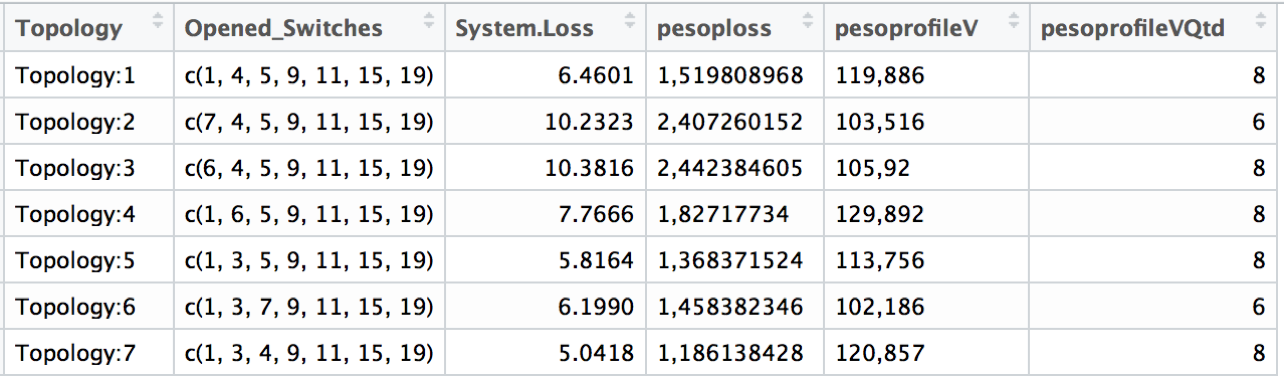}
\caption{Similar stored cases retrieved due to failures in bus 9 and 11}
\label{fig:CBR_Base_failure_9_11}
\end{figure}

The CBR module checks the existence of stored cases in its database with similar characteristics in relation to the current case with 92\% similarity with a computational time of 0.001 sec (Figure \ref{fig:CBR_Base_failure_9_11}).

	%=== Figura 9 ====

The framework is flexible and can be adjusted by the manager to look for similar cases according to established priorities.

\begin{figure}[ht]
\centering
\includegraphics[width=0.5\textwidth]{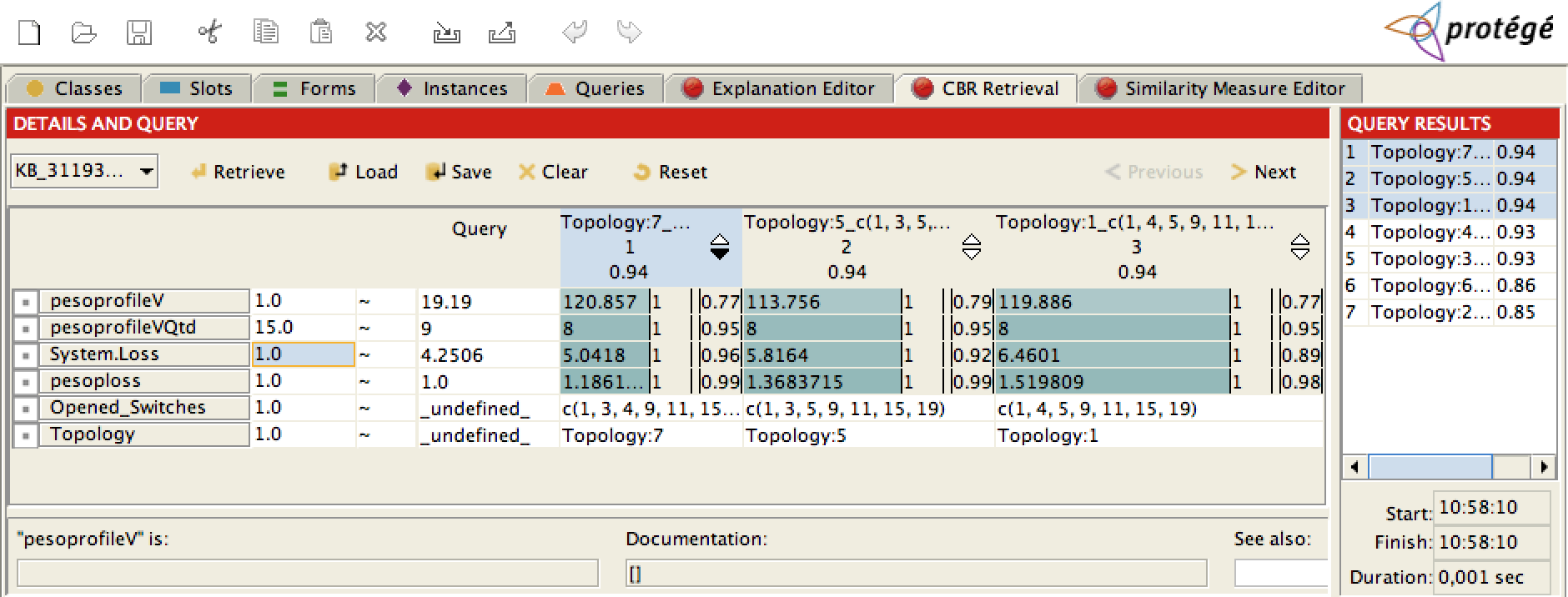}
\caption{Similar cases retrieved with priority.}
\label{fig:CBR_Base_failure_9_11_Similarity_pesoprofileVQtd.}
\end{figure}

If, for instance, the manager opted for cases giving priority to attribute "pesoprofileVQtd", the CBR module retrieves another set of cases up to 94\% of similarity as shown in Figure \ref{fig:CBR_Base_failure_9_11_Similarity_pesoprofileVQtd.}. In this simulation, three "best" cases have the same similarity. So the manager could choose the retrieve case combining priorities with more than one attributes to assist in the decision-making process. 

%%=== Figura 12 ====

In summary, the CBR module retrieves the \textit{n} cases that are maximally similar to the target problem in the database according to the policies and priorities determined by the network manager. 

The CBR-SGRec framework learning process occurs whenever its solves a problem successfully. In order to the system to keep up-to-date and continually evolve, the learning process remember this situation in the future as a new case. Its ability to solve problems improves as new cases are stored (learned) in its database. The CBR adaptation process in the reuse phase is done by evaluating the degree of initial match based on the similarity value. This is done using the general domain knowledge or requesting network manager directions. Adaptation is achieved by parameter adjustment defined by rules or configuration methods according to policies and constraints.

\section{Final Considerations}

This paper presented the CBR-SGRec framework that uses Case-Based Reasoning (CBR) coupled with the HATSGA algorithm for fast reconfiguration of large distribution power networks in the Smart Grid context.

The CBR-SGRec framework is cognitive and has the capability to compute new network reconfiguration topologies nearly on-the-fly.

The cognitive characteristic of CBR-SGRec is supported by the CBR module that, firstly, reduces the time required to compute new network configurations by finding out similar previously used solutions. Secondly, CBR module is capable to learn and incorporate new knowledge to its database by keeping new successful reconfiguration.

The on-the-fly characteristic of the CBR-SGRec framework depends of two components: the time required to compute a new reconfiguration by HATSGA algorithm and the time consumed to search for similar solutions at the database by the CBR module. HATSGA keeps the reconfiguration computing time to the minimal found in the literature. In addition to that, HATSGA has a nearly linear increase of its computational time with an experiential increase in network complexity (number of topologies). CBR module, as indicated in the proof-of-concept, consumes a much smaller amount of computing time to search similar solutions and, as such, does not compromise the on-the-fly characteristics of the framework.

In summary, the use of the CBR machine learning technique within the CBR-SGRec framework has  proven possible to have a  decision-making tool supporting the reconfiguration management process in the Smart Grid. This approach fundamentally reduces the human intervention for a rather complex management task requiring accurate engineering knowledge.

\bibliographystyle{IEEEtran}
\bibliography{biblio}
%\bibliography{sbc-template}

\end{document}